\documentclass[letterpaper, 10 pt, conference]{ieeeconf}  

\IEEEoverridecommandlockouts                              
\usepackage{graphics} 
\usepackage{epsfig} 
\usepackage{times} 
\usepackage{amsmath} 
\usepackage{amssymb}  
\usepackage{threeparttable}
\usepackage{multirow} 
\usepackage{xcolor} 
\usepackage{algorithm}
\usepackage{algpseudocode} 
\usepackage{booktabs}
\usepackage{lipsum} 

\title{\LARGE \bf
CL-CoTNav: Closed-Loop Hierarchical Chain-of-Thought for Zero-Shot Object-Goal Navigation with Vision-Language Models
}

\author{Yuxin Cai$^{1,2}$, 
Xiangkun He$^{1}$, 
Maonan Wang$^{3}$,
Hongliang Guo$^{2}$, 
Wei-Yun Yau$^{2}$, 
and Chen Lv$^{1,}$$^{\ast}$
\thanks{$^{1}$School of Mechanical and Aerospace Engineering, 
        Nanyang Technological University, Singapore
        {\tt\small \{caiy0039, 
        xiangkun.he, 
        lyuchen\}@ntu.edu.sg}}%
\thanks{$^{2}$Institute for Infocomm Research (I2R), 
        Agency for Science, Technology and Research (ASTAR), Singapore
        {\tt\small $\{$stucaiy, 
        guo$\_$hongliang, 
        wyyau$\}$@i2r.a-star.edu.sg}}%
\thanks{$^{3}$School of Science and Engineering, the Chinese University of Hong Kong, Shenzhen, China and the Shanghai AI Laboratory, Shanghai, China
{\tt\small $\{$maonanwang$\}$@link.cuhk.edu.cn}}%
\thanks{*Corresponding author}%
}

\begin{document}

\maketitle
\thispagestyle{empty}
\pagestyle{empty}

\begin{abstract}
Visual Object Goal Navigation (ObjectNav) requires a robot to locate a target object in an unseen environment using egocentric observations.
However, decision-making policies often struggle to transfer to unseen environments and novel target objects, which is the core generalization problem. Traditional end-to-end learning methods exacerbate this issue, as they rely on memorizing spatial patterns rather than employing structured reasoning, limiting their ability to generalize effectively.
In this letter, we introduce Closed-Loop Hierarchical Chain-of-Thought Navigation (CL-CoTNav), a vision-language model (VLM)-driven ObjectNav framework that integrates structured reasoning and closed-loop feedback into navigation decision-making.
To enhance generalization, we fine-tune a VLM using multi-turn question-answering (QA) data derived from human demonstration trajectories. This structured dataset enables hierarchical Chain-of-Thought (H-CoT) prompting, systematically extracting compositional knowledge to refine perception and decision-making, inspired by the human cognitive process of locating a target object through iterative reasoning steps.
Additionally, we propose a Closed-Loop H-CoT mechanism that incorporates detection and reasoning confidence scores into training. This adaptive weighting strategy guides the model to prioritize high-confidence data pairs, mitigating the impact of noisy inputs and enhancing robustness against hallucinated or incorrect reasoning.
Extensive experiments in the AI Habitat environment demonstrate CL-CoTNav's superior generalization to unseen scenes and novel object categories. Our method consistently outperforms state-of-the-art approaches in navigation success rate (SR) and success weighted by path length (SPL) by 22.4\%. We release our datasets, models, and supplementary videos on our project page.
\end{abstract}

\vspace{2mm}

\section{Introduction} 
\begin{figure}[h]
    \centering 
    \includegraphics[width=0.49\textwidth]{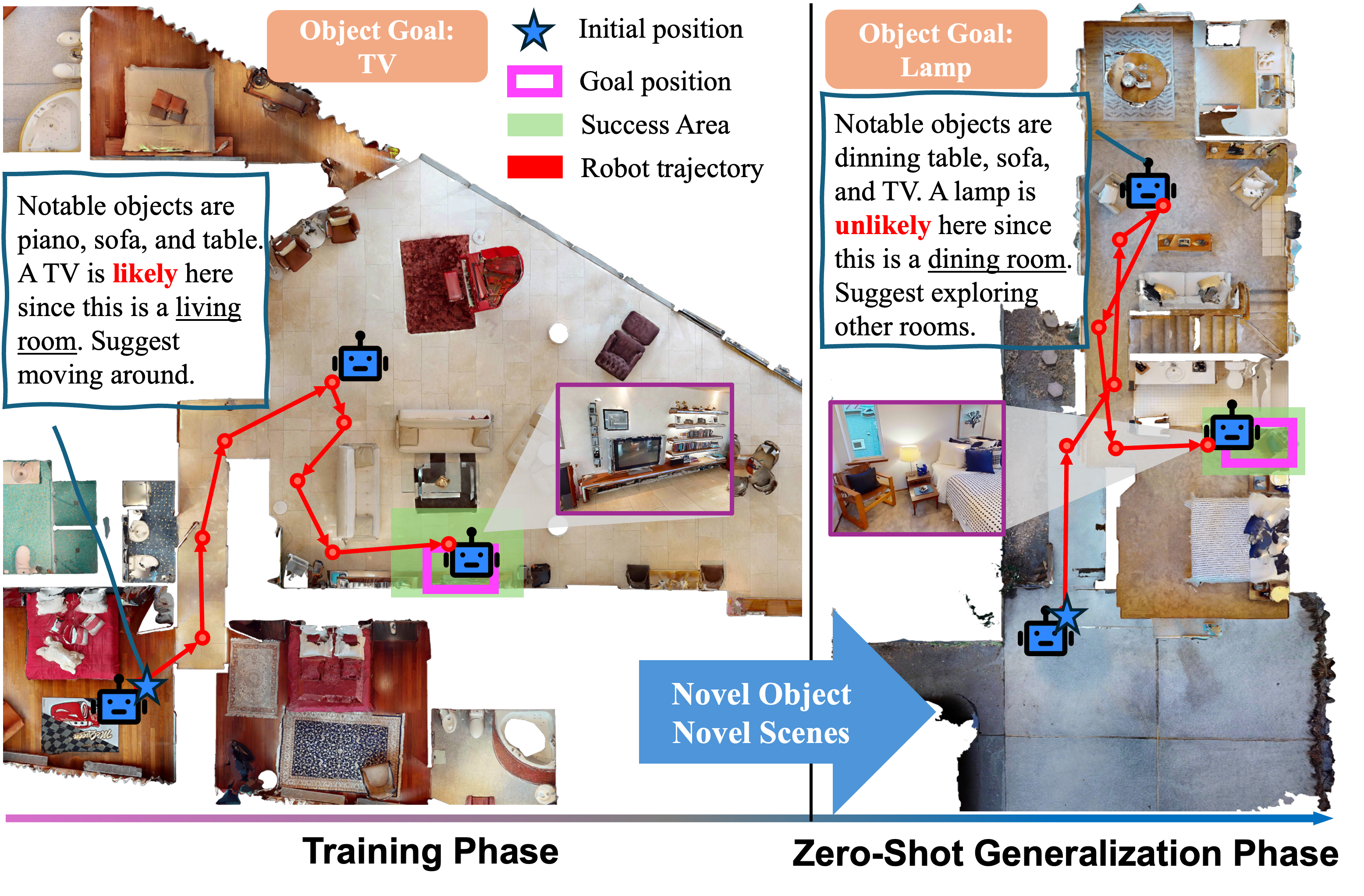}
    \caption{Overview of Zero-Shot Object Navigation (ZSON). The left side illustrates the training phase, where an robot trains to navigate within seen scenes sets and towards target objects sets. The right side represents the zero-shot generalization phase, where the trained policy is evaluated in unseen target objects and novel scenes without further training. The figure also highlights the proposed H-CoT process, where it reasons about likely and unlikely object locations based on object-object and object-scene relationships. 
    }

    \label{fig:problem}
\end{figure}

Humans efficiently navigate unfamiliar environments to find an object by reasoning semantic relationships \cite{yang2018visual}—for instance, kitchens are typically adjacent to living rooms, and exit signs indicate pathways to an exit. This structured reasoning allows humans to infer the probable location of a target object without exhaustive exploration or explicit SLAM-based mapping. Replicating this capability in robots is fundamental to Object Goal Navigation (ObjectNav), where a robot must learn to locate an instance of a specified object category in an unseen environment using only egocentric observations \cite{sun2024survey}. However, achieving such reasoning-driven navigation remains an open challenge, particularly when addressing zero-shot generalization, where robot must navigate to previously unseen objects or adapt to novel scene layouts without retraining, making Zero-Shot Object Navigation (ZSON) a critical yet unsolved problem \cite{majumdar2022zson}, as shown in Fig.\ref{fig:problem}.

The ZSON problem presents two primary challenges \cite{zhao2023semantic}. First, unseen object generalization requires robot to infer the probable location of novel target objects based on learned semantic relationships rather than memorizing specific instances. For example, if robot has learned to find a "chair," it should be able to infer the likely placement of a "stool" based on shared spatial and functional attributes, rather than requiring explicit training on stools. Second, unseen scene generalization requires robot to learn layout patterns rather than memorizing specific room arrangements. Real-world indoor spaces exhibit significant variations in spatial arrangements but share same layout pattern. Effective generalization demands the ability to infer high-level spatial relationships—such as recognizing that a kitchen is typically adjacent to a dining area—rather than relying on rigid spatial distributions observed during training.
Addressing these challenges necessitates structured semantic reasoning beyond conventional feature-based learning. A robust ZSON policy should explicitly incorporate object-object and object-scene co-occurrence relationships to improve generalization across unseen objects and scenes while maintaining efficient navigation.

Existing approaches to ZSON primarily fall into two paradigms: modular pipelines and end-to-end reinforcement learning (RL) \cite{gervet2023navigating}. Modular pipelines decompose navigation into semantic mapping, frontier goal selection, and motion planning, explicitly constructing top-down representations to infer potential object locations. 
While interpretable by map,
these methods heavily depend on accurate localization and mapping, making them sensitive to errors in depth sensing. This poses a significant challenge for real-world deployment, where sensor noise and dynamic scene variations degrade decision accuracy.
In contrast, end-to-end RL methods directly optimize policies from egocentric visual inputs to output actions \cite{ramrakhya2022habitat}. 
While capable of capturing complex visual-action associations, these approaches require extensive interaction data for training.
Both paradigms also share one crucial limitation: they lack a principled mechanism to integrate semantic reasoning and uncertainty estimation in-the-loop, limiting their generalization ability.

Recent advances in VLMs have introduced new opportunities for incorporating commonsense reasoning priors into navigation \cite{shah2023lm, yu2023l3mvn}. These models excel at object recognition and reasoning, making them valuable for introducing semantic priors and inferring semantic relationships \cite{zhang2024vision, wen2025zero}. 
However, directly applying such models to ObjectNav task remains challenging \cite{guo2022domain}. While language models can generate high-level narratives about reaching a target, they lack direct grounding in real-world observations and often produce hallucinated or semantically incorrect reasoning. 
Existing vision-language navigation methods also either lack structured intermediate reasoning steps or fail to incorporate incorrect reasoning, leading to unreliable output.

To address these limitations, we propose a VLM-driven navigation framework that introduces a hierarchical chain-of-thought reasoning process. Instead of directly predicting actions, our method decomposes ObjectNav into structured multi-turn question-answering steps that explicitly model human-like decision-making, inspired by how humans iteratively refine their understanding of the environment.
In addition, to mitigate noisy RGB observations or incorrect commonsense reasoning associations, we introduce a closed-loop mechanism that incorporates confidence-weighted adaptive learning into training, prioritizing high-confidence trajectories while reducing the influence of unreliable predictions. This refinement improves generalization to unseen objects and environments while ensuring robust navigation decisions.
Our contributions can be summarized as follows:
\begin{itemize}
    \item We introduce CL-CoTNav, a VLM-driven ObjectNav framework that integrates structured reasoning into navigation decision-making. We fine-tune small-scale VLM using multi-turn question-answering (QA) data derived from human demonstration trajectories. This structured dataset enables hierarchical Chain-of-Thought (H-CoT) prompting, iteratively extracting compositional knowledge from pretrained foundation models, serve as auxiliary navigation guidance labels.
    \item We further develop a closed-loop H-CoT mechanism that introduces confidence scores into training through adaptive loss weighting, reducing the impact of both noisy RGB observations and unreliable semantic reasoning associations.
    \item We conduct extensive experiments in AI Habitat, demonstrating that our method outperforms state-of-the-art ObjectNav baselines in success rate and success weighted by path length. Our results highlight the benefits of structured reasoning and confidence-aware learning for zero-shot object navigation.
\end{itemize}

\section{Related Work}

\subsection{Object Goal Visual Navigation} 
Object Goal Visual Navigation (ObjectNav) requires an agent to locate an instance of a target object category in an unseen environment using only visual inputs \cite{sun2024survey}. Existing approaches can be categorized into modular learning and end-to-end learning \cite{gervet2023navigating}.  
Modular learning methods decompose the navigation task into semantic mapping, goal selection, and motion planning. These methods construct a top-down semantic map, select exploration goals based on learned or heuristic policies, and execute low-level actions through a local planner \cite{gervet2023navigating}. Representative pipelines \cite{chaplot2020object, zhang2018semantic} incrementally build episodic semantic maps while employing goal-agnostic, frontier-based exploration. The semantic exploration policy determines navigation goals by leveraging learned priors on object spatial relationships, while the local planner \cite{sethian1996fast} generates paths and executes actions. The exploration policy operates at a coarse time scale, whereas the planner continuously updates the map and refines the path at a finer scale. Although modular approaches are interpretable and transferable to real-world settings, their reliance on accurate localization and mapping limits practice in large-scale environments.  
End-to-end learning methods, in contrast, directly map raw sensor inputs and goal descriptions to navigation actions using deep neural networks, bypassing explicit mapping. 
It learns implicit representations of the observation before inputting it into the navigation policy, exploiting the object relationships or semantic contexts, aiming for a more robust navigation policy.
\cite{lian2024tdanet} incorporate attention mechanisms to prioritize relevant observed objects. 
Meta-learning strategies \cite{wortsman2019learning} enable adaptive navigation in unseen environments without explicit supervision, and graph neural networks \cite{chen2024socially} have been employed to model object relationships and extract semantic interaction features. 
\cite{cai2024bridging} introduce an image-level representation to bridge the gap between RGB inputs and control-space actions. 
While end-to-end approaches captures complex visual-action associations, they require large-scale interactive training data and often struggle with long-horizon dependencies and memory retention.  

In this work, we adopt this end-to-end learning paradigm, integrating semantic relationships reasoning between the target object, observed objects and scene. Our approach is designed to be generalizable across unseen scenes and object categories, enhancing adaptability in real-world ObjectNav scenarios.

\subsection{Zero-Shot Object Goal Navigation} 
Traditional ObjectNav methods are trained on a fixed set of object categories and scenes \cite{zhu2017target, yang2018visual}, limiting their ability to generalize beyond seen cases. In contrast, humans can effortlessly locate novel objects in unfamiliar environments without prior exposure. Achieving similar generalization in ObjectNav requires disentangling navigation ability from specific training scenes and target objects. To evaluate this, experiments typically partition object target classes and scenes into seen and unseen categories, assessing the model’s capability to navigate to novel objects and scenes without additional training.  
End-to-end methods such as SSNet \cite{zhao2023zero} incorporate object detection results and cosine similarity between word embeddings to prevent class-specific policy overfitting. SPNet \cite{zhao2023semantic} refines policy learning through object-goal embeddings that guide action selection based on semantic similarity. EmbCLIP \cite{khandelwal2022simple} and ZSON \cite{majumdar2022zson} utilize pretrained vision encoders and text-based embeddings \cite{radford2021learning} to establish semantic relationships between target objects and observed scenes, improving generalization without requiring additional annotated training data.  

Despite advances in semantic embeddings and target-guided exploration, ZSON remains challenging due to the variability of object and scene distributions and the absence of unified features and prior knowledge for efficient search. To address this, our method leverages human-demonstrated trajectories, interpreting them hierarchically to integrate structured priors, semantic reasoning, and agnostic exploration, thereby achieving robust generalization without extensive retraining.

\begin{figure*}[h]
    \centering 
    \includegraphics[width=1\textwidth]{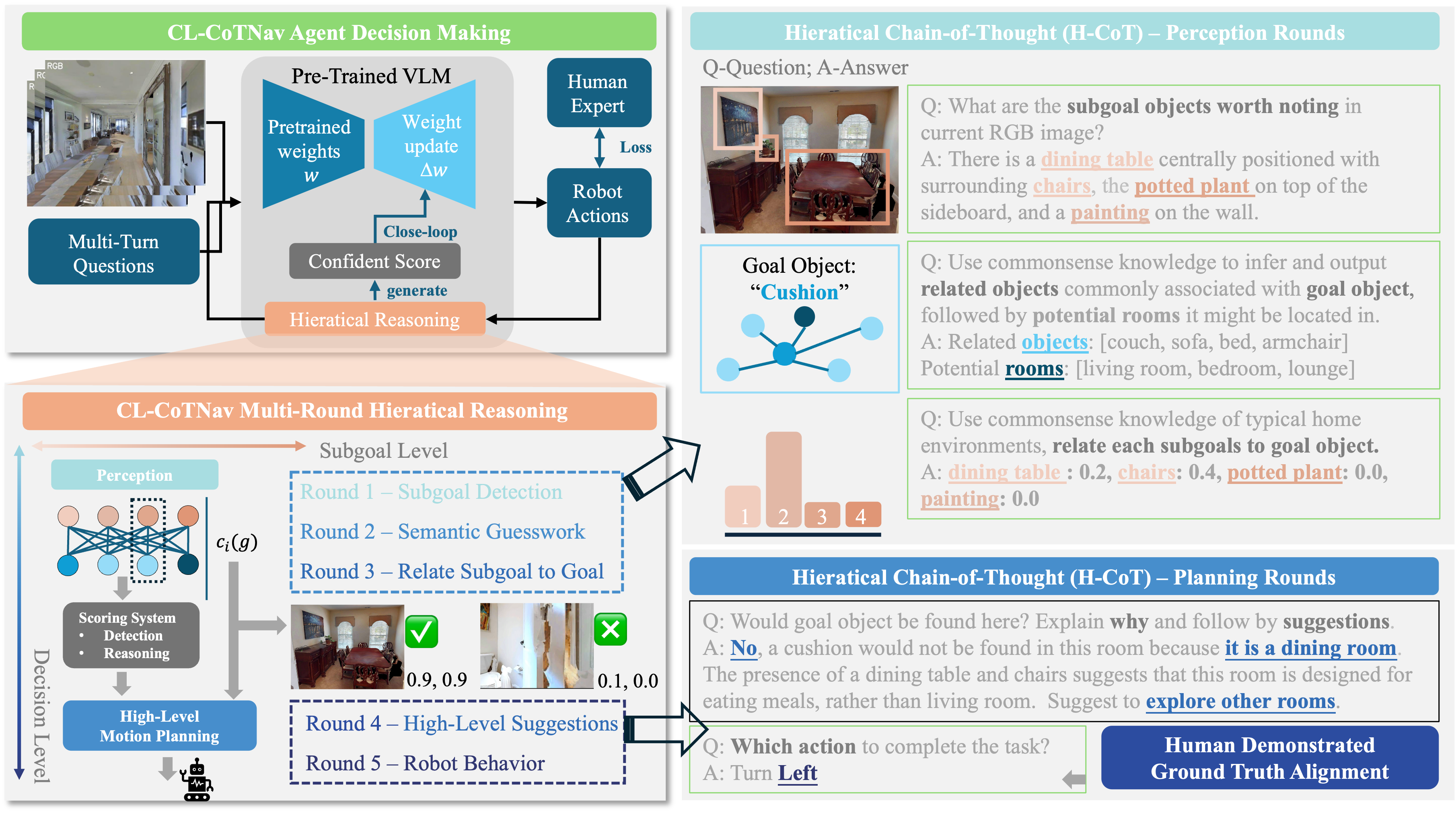}
    \caption{Overview of CL-CoTNav. We finetune VLM using multi-turn QA data derived from human demonstration trajectories. This structured dataset enables H-CoT prompting, including two main turns: perception and planning, to
    iteratively extract compositional knowledge from egocentric RGB observations through a sequence of large pre-trained models and finally aligned with human demonstration actions.
    A confidence scoring system is also generated to evaluate the reliability of each detection and reasoning step, which guide adaptive loss weighting during finetuning to improve robustness against noisy supervision.}
    \label{fig:network architecture}
\end{figure*}

\subsection{Foundation Models for Visual Navigation}
Recent works integrate pre-trained multimodal models into ObjectNav, reducing the need for training from scratch while leveraging strong visual recognition or reasoning. These methods primarily use foundation models for (1) exploration guidance by relating observations to target objects or (2) policy learning for direct action prediction.  
For exploration guidance, 
VLFM \cite{yokoyama2024vlfm} employs BLIP-2 \cite{li2023blip} to compute cosine similarity between observations and target prompts, projecting scores onto a semantic map to guide frontier-based exploration. 
OpenFMNav \cite{kuang2024openfmnav} decomposes ObjectNav into sequential stages, utilizing vision-language models (VLMs) for perception, reasoning, and constructing a semantic score map for language-guided navigation. 
CoW \cite{li2023blip} provides object grounding, similarly projecting relevance scores onto a top-down map. 
While these methods enable zero-shot navigation without additional training, frozen VLMs may produce incorrect associations due to biases or hallucinations, lacking mechanisms for real-time correction.  
For policy learning, 
ViNT \cite{shah2023vint} encodes vision-language features into a transformer-based policy, requiring an additional adapter to map from token space to action space. 
DivScene \cite{wang2024divscene} directly fine-tunes VLMs using imitation learning, but its reliance on annotated shortest paths—assuming prior knowledge of the target object’s location—limits the model’s ability to learn efficient search strategies.  

To address these limitations, our method directly employs a compact VLM as the full navigation policy, introducing a structured reasoning framework that extracts compositional knowledge from human-demonstrated trajectories. Additionally, a closed-loop feedback mechanism mitigates VLM errors, enhancing adaptability and robustness in unseen environments.

\section{Methodology}

\subsection{Problem formulation}
In the ObjectNav task, consider a set of seen object goal classes \( C_{\text{train}} = \{c_1, c_2, \dots, c_n\} \) available during training, where \( n \) is the total number of seen classes, and a set of seen scenes \( X_{\text{train}} \). During zero-shot testing, the agent navigates to objects from an unseen target class set \( C_{\text{test}} = \{u_1, u_2, \dots, u_m\} \), where \( m \) is the total number of unseen classes, within unseen scenes \( X_{\text{test}} \). The training and testing sets are disjoint, i.e., \( X_{\text{train}} \cap X_{\text{test}} = \emptyset \) and \( C_{\text{train}} \cap C_{\text{test}} = \emptyset \). The robot is first trained to navigate to a target object \( c_i \) from \( C_{\text{train}} \) within a scene \( X_{\text{train}, i} \), given egocentric RGB observations, and is later evaluated in a zero-shot setting where it must navigate to a target object from \( C_{\text{test}} \) in an unseen scene \( X_{\text{test}, i} \). Each episode initializes the agent at a random position \( p_0^i \) with a random orientation in scene \( s_0^i \), and the target object category \( c_i \) is provided. An episode is therefore characterized by \( T_i = \{s_i, c_i, p_i, o_i, a_i\} \). At each time step \( t \), the agent receives an observation \( o_t^i \) from its current viewpoint and selects an action \( a_t^i \). The observation consists of an RGB image, the agent’s location and orientation, and the target object category. The action space \( A \) consists of six discrete actions: \textit{move\_forward}, which moves the agent forward by 25 cm; \textit{turn\_left} and \textit{turn\_right}, which rotate the agent 30° left or right; \textit{look\_up} and \textit{look\_down}, which adjust the camera pitch by 30°; and \textit{stop}, which signals that the target has been reached. An episode is considered successful if the agent executes the \textit{stop} action and the target object is visible as well as within 0.2 m of the target. Each episode is constrained to a maximum of 500 time steps.

\subsection{Overview}

The CL-CoTNav framework is illustrated in Fig.~\ref{fig:network architecture}. To model human-like reasoning in ObjectNav, our method transforms egocentric RGB observations and corresponding human demonstration trajectories into structured, multi-turn question-answering (QA) sequences. This is achieved using a pipeline of pre-trained large vision-language and language models. Each QA sequence aligns intermediate reasoning steps with ground-truth actions, enabling supervision beyond low-level behavior cloning.

At the core of the framework is the H-CoT prompting process, which decomposes ObjectNav into two reasoning phases: Perception Rounds and Planning Rounds. In the Perception Rounds, the system identifies salient objects from the current RGB observation, infers semantic co-occurrence relationships at both the room and object levels, and constructs a structured scene-level context. These steps allow the model to hypothesize where the target object is likely to be found, mimicking the cognitive strategies used by humans during navigation. The Planning Rounds then reason over the accumulated context, producing high-level navigation suggestions (e.g., "turn left" or "explore another room") that are discretized into executable control actions. This hierarchical prompting process acts as an interpretable bridge between perception and control, promoting generalization between perception and control.

To enhance robustness and learning efficiency, we further propose the Closed-Loop H-CoT mechanism. While H-CoT provides structured supervision, it remains vulnerable to label errors due to noisy visual inputs or incorrect semantic reasoning associations. To mitigate this, we introduce a confidence scoring system that evaluates the reliability of each QA pair based on reasoning consistency and input quality. During training, these confidence scores are integrated into an adaptive loss function that modulates the influence of each sample. High-confidence samples contribute more strongly, while uncertain or potentially hallucinated reasoning paths are down-weighted.

The complete QA dataset, annotated with hierarchical reasoning steps and associated confidence scores, is then used to fine-tune a compact vision-language model via Low-Rank Adaptation (LoRA). By combining hierarchical prompting with closed-loop learning, CL-CoTNav achieves robust generalization to unseen scenes and novel object categories.
We next detail the two key components of our approach: the H-CoT reasoning process and the Closed-Loop confidence integration mechanism.

\subsection{Hierarchical Chain-of-Thought (H-CoT)}

To enrich the supervision signal beyond discrete human-labeled actions, we introduce the Hierarchical Chain-of-Thought (H-CoT) framework—a structured reasoning process that mimics human cognitive strategies for navigation. H-CoT decomposes the ObjectNav task into two multi-turn reasoning stages: Perception Rounds for semantic scene understanding and Planning Rounds for informed decision-making. This design enables compositional reasoning over spatial and semantic cues, allowing the model to generalize beyond training distributions.

In the \textbf{Perception Rounds}, a sequence of QA steps is applied to egocentric RGB observations to extract structured scene context. The first round identifies subgoal objects and their spatial arrangements (e.g., ``a dining table is centrally positioned with surrounding chair"), capturing the visual layout of the environment. Subsequent rounds perform semantic guesswork through two levels of association. At the room level, the model infers the scene type based on typical object co-occurrence (e.g., ``a TV and sofa suggest a living room"). At the object level, it relates detected subgoal objects to the target via commonsense priors (e.g., ``a cushion is likely near a couch, but not near a stove"). These associations allow the model to hypothesize the target’s likely presence without direct observation.
Each subgoal object is assigned a relevance score that reflects its semantic proximity to the target object, serving as a soft attention mechanism. This scoring filters out spurious detections and prioritizes contextual cues that are semantically informative, forming a rich representation of the current scene to guide object navigation.

Building on this context, the \textbf{Planning Rounds} generate navigation suggestions grounded in the inferred scene semantics. The model evaluates whether the current room is a plausible location for the target; if not, it recommends exploration strategies such as ``explore another room" or ``turn around." These suggestions are then mapped into high-level textual decisions, which are further discretized into text-based executable control actions. Critically, these planning outputs are aligned with human demonstration actions, grounding abstract reasoning in behaviorally relevant supervision.

To generate this hierarchical supervision, we annotate human demonstration data using a pipeline of pre-trained vision-language models, resulting in this dataset of structured multi-turn QA pairs aligned with human actions. This dataset enables supervised training that incorporates both detection, reasoning and control, bridging the gap between low-level action imitation and high-level semantic understanding.
By explicitly modeling semantic co-occurrence relationships and decomposing decision-making into interpretable stages, H-CoT provides a strong inductive bias for zero-shot generalization, allows the agent to reason compositionally about new objects and scenes.

\subsection{Closed-Loop H-CoT Mechanism}

While the H-CoT framework introduces structured reasoning for ObjectNav, it remains vulnerable to failures caused by noisy visual inputs, hallucinated associations, or unreliable outputs from the underlying vision-language models. These detection and reasoning inconsistencies can degrade navigation performance when treated equally during training. To address this, we propose the Closed-Loop H-CoT Mechanism, a feedback-driven strategy that introduces reasoning confidence scores to modulate training dynamics.

In standard imitation learning, all training samples are treated with equal importance, regardless of their semantic clarity or visual quality. In contrast, our approach attaches a confidence score to each sample during H-CoT generation, capturing the reliability of both detection (object grounding) and reasoning (semantic inference) at each turn. Specifically, for each multi-turn QA sequence labeled by pre-trained models, we parse out the final text-based action suggestion and compare it to the human-demonstrated ground truth action. The degree of semantic alignment—combined with visual detection certainty—forms a confidence score \( c_i \in [0,1] \) for each reasoning trajectory.

These confidence scores are then integrated into training via an adaptive loss weighting mechanism. ObjectNav is formulated as a multi-class classification task, where the model predicts a discrete navigation action (e.g., \textit{forward}, \textit{left}, \textit{right}) conditioned on the RGB image and the corresponding structured QA prompt. The baseline training objective is defined using categorical cross-entropy:
\begin{equation}
    L_{\text{CE}} = - \log \hat{y}_{i, y_i}
\end{equation}
where \( \hat{y}_{i, y_i} \) is the predicted probability assigned to the correct action label \( y_i \).
To prioritize learning from trustworthy trajectories and downweight unreliable supervision, we define a sigmoid-based adaptive loss function:
\begin{equation}
    L_{\text{adaptive}} = \frac{1}{1 + \exp(-\alpha (c_i - \beta))} \cdot (- \log \hat{y}_{i, y_i})
\end{equation}
Here, \( c_i \) is the confidence score of the \( i \)-th sample, while \( \alpha \) and \( \beta \) are hyperparameters controlling the sharpness and threshold of the weighting function.

This closed-loop design enables the model to selectively attend to high-quality detection and reasoning timesteps while minimizing the impact of noise or hallucinations from photorealistic scene and language model. By integrating confidence-weighted fine-tuning into the training loop, the Closed-Loop H-CoT mechanism improves generalization to unseen environments and enhances the robustness of decision-making under real-world visual uncertainty.

\section{Experiment}

In this section, we first describe the details of our experiment settings, including the dataset split, simulation platform, and training parameter. Then, we evaluate our method and other ZSON
methods through the commonly used Habitat platform and discuss the experimental results.

\subsection{Dataset}
We evaluate our method using the Matterport3D (MP3D) environment \cite{chang2017matterport3d} in the Habitat simulator \cite{puig2023habitat}, which provides high-resolution, photo-realistic indoor scenes with 21 object goal categories. Our experiments follow the standard Zero-Shot Object Navigation setting, which evaluates generalization across both novel object categories and unseen scenes.
Training is conducted on the MP3D-HD-70k dataset \cite{ramrakhya2022habitat}, which contains over 70,000 human demonstration trajectories collected across 56 scenes. To ensure data quality, we remove failed trajectories, filter non-navigable starting positions, and cap all episodes to a maximum length of 500 steps. After preprocessing, we obtain cleaned subsets of 35k, 50k, and 70k demonstrations, balanced across object classes and scene types.
We design two evaluation protocols to measure generalization, as shown in Table \ref{table:dataset split}: (1) object generalization, where the target categories differ between training and test, and (2) scene generalization, where the environments differ.

\begin{table}[h]\centering
\caption{Breakdown of train and test datasets for scene and object generalization experiments. Inside () indicate episode or target numbers after cleaning.}
\begin{tabular}{@{}lcccc@{}}
\hline\hline
\vspace{0.2mm} \\
\textbf{Split} & \textbf{Dataset}       & \textbf{Scenes} & \textbf{Episodes} & \textbf{Targets} \\
\midrule
\multicolumn{5}{c}{\textbf{Scene Generalization}} \\ \midrule
Train   & MP3D-HD-70k      & 56           & 70,176 (53,827)   & 28 (21) \\
Train   & MP3D-HD-50k      & 40           & 49,778 (37,925)   & 28 (21) \\
Train   & MP3D-HD-35k      & 28           & 34,641 (26,517)   & 28 (21) \\
Test    & MP3D-Val         & 11           & 2,195 (1,148)     & 21      \\  
\midrule
\multicolumn{5}{c}{\textbf{Object Generalization}} \\ \midrule
Train   & MP3D-HD-35k-C16  & 28           & 20,595            & 16      \\  
Test    & MP3D-HD-35k-C05  & 28           & 5,922             & 5       \\  
\hline\hline
\label{table:dataset split}
\end{tabular}
\end{table}

\textbf{Object Generalization.}  
For this split, we adopt the setting from \cite{zhao2023semantic}, where the 21 object categories in MP3D are divided into 16 seen and 5 unseen classes. The training set (MP3D-HD-35k-C16) includes trajectories involving only the seen classes, while the test set (MP3D-HD-35k-C05) uses the same 28 training scenes but targets the five unseen categories: \textit{counter, bed, toilet, chest\_of\_drawers, plant}. This setting evaluates the model's ability to reason over novel object semantics not encountered during training. The remaining 16 categories—e.g., \textit{chair, table, sofa, tv\_monitor, sink}—are exclusively used for training.

\textbf{Scene Generalization.}  
To assess generalization to unseen spatial layouts, we train on the full MP3D-HD dataset using subsets of 28, 40, or 56 scenes (i.e., MP3D-HD-35k/50k/70k), each containing trajectories across all 21 object categories. Evaluation is performed on the MP3D-Val set, which includes 2,195 episodes across 11 held-out scenes that do not overlap with any training environments. This setting focuses on the model’s ability to adapt to novel layouts and scene compositions, even when the object categories remain the same.
Table~\ref{table:dataset split} summarizes the number of scenes, episodes, and target categories used across all training and evaluation configurations.

\subsection{Implementation Details}

In the ObjectNav task, the agent is required to search for an instance of a specified object category (e.g., \textit{bed}) within an unseen environment using only egocentric perception. The agent is equipped with an RGB camera, a depth sensor, and an odometry sensor that provides its pose relative to the episode's starting position.
The simulated robot is 0.88 meters tall with a 0.18-meter radius. It captures $480 \times 640$ RGB-D observations through a forward-facing camera mounted at a height of 0.88 meters, with a horizontal field of view (HFoV) of 79 degrees. All experiments are conducted using the Habitat-Lab simulation platform \cite{puig2023habitat}, and models are implemented in PyTorch \cite{paszke2019pytorch}.

To generate multi-round QA annotations for our dataset, we employ the following pre-trained models for different submodules: Qwen-VL-Chat \cite{Qwen-VL} for subgoal detection, Qwen-7B \cite{qwen} for semantic guesswork and object-target association, and ChatGPT-3.5-turbo for high-level action suggestion and scene-level reasoning. Example of conversation template can be found in Fig.\ref{fig:network architecture}.
For navigation policy learning, we fine-tune based on InternVL2 \cite{chen2024internvl} framework, elaborate a 2B-parameter vision-language model, using the LoRA \cite{hu2022lora} technique. Fine-tuning is performed on a compute node equipped with 4 NVIDIA V100 GPUs. We use a batch size of 16 and train for 3 epochs, which takes approximately 19 hours on the MP3D-HD-50k dataset. 
The LoRA-specific fine-tuning hyperparameters are summarized in Table~\ref{tab:lora_hyperparams}.

\begin{table}[t]
    \centering
    \caption{Hyperparameters for LoRA Fine-tuning}
    \begin{tabular}{l c}
        \hline \hline
        \textbf{Parameter} & \textbf{Value} \\ 
        \hline
        LoRA rank ($r$) & 8 \\ 
        LoRA scaling factor ($\alpha$) & 16 \\ 
        LoRA dropout & 0.05 \\ 
        Learning rate & $3 \times 10^{-4}$ \\ 
        Batch size & 16 \\ 
        Gradient accumulation steps & 4 \\ 
        Weight decay & 0.006 \\ 
        Warmup steps & 500 \\ 
        \hline \hline
    \end{tabular}
    \label{tab:lora_hyperparams}
\end{table}

\subsection{Metrics}
We follow \cite{anderson2018evaluation} to evaluate our method using Success Rate (SR),  
Success Weighted by Path Length (SPL), and Soft SPL for object-goal navigation tasks.  
SR is defined as: $\frac{1}{N} \sum_{i=1}^N S_i$
where $S_i = 1$ if the robot successfully reaches the target; otherwise, the episode is considered a failure.  
Success Weighted by Path Length (SPL) is defined as:  
$SPL = \frac{1}{N} \sum_{i=1}^N \frac{l_i}{\max(l_i, p_i)}$
where $l_i$ is the shortest path from the start position to a successful stop position, and $p_i$ is the robot's actual trajectory length in episode \( i \).  
Finally, Soft SPL \cite{datta2021integrating} accounts for navigation efficiency while incorporating partial progress toward the goal.  

\subsection{Comparison Models}

We compare our proposed method with several representative baseline and state-of-the-art (SOTA) approaches in ObjectNav, spanning RL, imitation learning (IL), and VLM paradigms.

\textbf{Baseline}~\cite{zhu2017target}: A RL approach trained from scratch using egocentric RGB inputs. It directly use a pre-trained ResNet to extract a
1-D visual feature from the RGB observation and concatenate with the semantic embedding of the target class as the input of the policy network.
The policy is learned end-to-end using PPO without incorporating semantic priors.

\textbf{Habitat-Web}~\cite{ramrakhya2022habitat}: An IL method that trains ObjectNav agents directly from human demonstration trajectories. The model maps egocentric RGB observations to expert-labeled actions via an end-to-end MLP policy, similar to Baseline.

\textbf{VLFM}~\cite{yokoyama2024vlfm}: A VLM-based modular navigation framework that employs BLIP-2~\cite{li2023blip} for semantic matching. It computes cosine similarity between the agent's current RGB view and the target object description, projecting scores onto a semantic map for goal selection. The model is frozen during deployment and not fine-tuned for ObjectNav tasks.

\textbf{SSNet}~\cite{zhao2023zero}: A RL-based zero-shot ObjectNav model that integrates object 
detection scores and word embedding similarity as anxillary input to policy learning. 

\textbf{DivScene}~\cite{wang2024divscene}: An VLM approach that also introduces chain-of-thought (CoT) reasoning for decision-making. Unlike our method, which learns an end-to-end policy from human behavior using fine-tuned VLMs, DivScene employs CoT supervision based on shortest path trajectories.

\subsection{Training Results}

Table~\ref{tab:train_results} presents the evaluation performance of CL-CoTNav and other baseline models on the training splits after training. All methods achieve comparable success rates (SR) and success weighted by path length (SPL), indicating that ObjectNav is generally learnable across different paradigms, including reinforcement learning, imitation learning, and VLM-based methods.

CL-CoTNav consistently achieves the highest SPL and competitive SR, demonstrating the effectiveness of structured hierarchical reasoning and confidence-aware fine-tuning. Compared to imitation learning approaches such as Habitat-Web and DivScene, CL-CoTNav exhibits notably higher SPL, suggesting that structured multi-round reasoning contributes to more efficient and purposeful trajectories.
While VLFM leverages frozen vision-language embeddings and commonsense priors, it performs worse than fine-tuned approaches, particularly in SPL. This confirms that adaptation to navigation-specific tasks is critical for fully exploiting the semantic reasoning capacity of large-scale vision-language models. Although DivScene incorporates chain-of-thought supervision into its imitation learning pipeline, its reliance on shortest-path ground truth limits its adaptability. In contrast, CL-CoTNav further benefits from closed-loop learning, improving robustness by emphasizing high-confidence reasoning during training.

We also observe that increasing the number of human demonstration trajectories improves performance, though with diminishing returns. Moving from 35k to 50k demonstrations yields substantial gains in both SR and SPL, while the improvement from 50k to 70k is more modest. This suggests that while imitation learning benefits from more data, its effectiveness saturates without structured supervision. In this context, our results highlight the importance of compositional reasoning over pure data scaling for achieving high-quality navigation performance.

\begin{table}[h]
\centering
\caption{Training Results for ObjectNav. We report Success Rate (SR) and Success weighted by Path Length (SPL) for each SOTA.}
\begin{tabular}{lccc}
\hline\hline
\textbf{Method} & \textbf{SR (\%)} & \textbf{SPL (\%)} & \textbf{Training} \\
\hline\hline
\multicolumn{4}{c}{\textbf{Object Goals}} \\
Baseline~\cite{zhu2017target} & 63.3 & 0.21 & Yes \\
SSNet~\cite{zhao2023zero} & 65.4 & 0.23 & Yes \\
Habitat-Web~\cite{ramrakhya2022habitat} & 69.1 & 0.26 & Yes \\
VLFM~\cite{yokoyama2024vlfm} & 70.1 & 0.28 & No \\
DivScene~\cite{wang2024divscene} & 73.8 & 0.30 & Yes \\
CL-CoTNav (MP3D-HD-35k-C16) & \textbf{74.1} & \textbf{0.31} & Yes \\
\hline\hline
\multicolumn{4}{c}{\textbf{Scenes}} \\
Baseline~\cite{zhu2017target} & 64.1 & 0.22 & Yes \\
SSNet~\cite{zhao2023zero} & 67.5 & 0.25 & Yes \\
Habitat-Web~\cite{ramrakhya2022habitat} & 70.8 & 0.27 & Yes \\
VLFM~\cite{yokoyama2024vlfm} & 71.3 & 0.28 & No \\
DivScene~\cite{wang2024divscene} & 75.6 & 0.32 & Yes \\
CL-CoTNav (MP3D-HD-35k) & 73.5 & 0.32 & Yes \\
CL-CoTNav (MP3D-HD-50k) & 74.8 & 0.35 & Yes \\
CL-CoTNav (MP3D-HD-70k) & \textbf{76.2} & \textbf{0.38} & Yes \\
\hline\hline
Humans & 93.7 & 42.5 \\
\hline\hline
\end{tabular}
\label{tab:train_results}
\end{table}

\begin{figure*}[h]
    \centering 
    \includegraphics[width=1\textwidth]{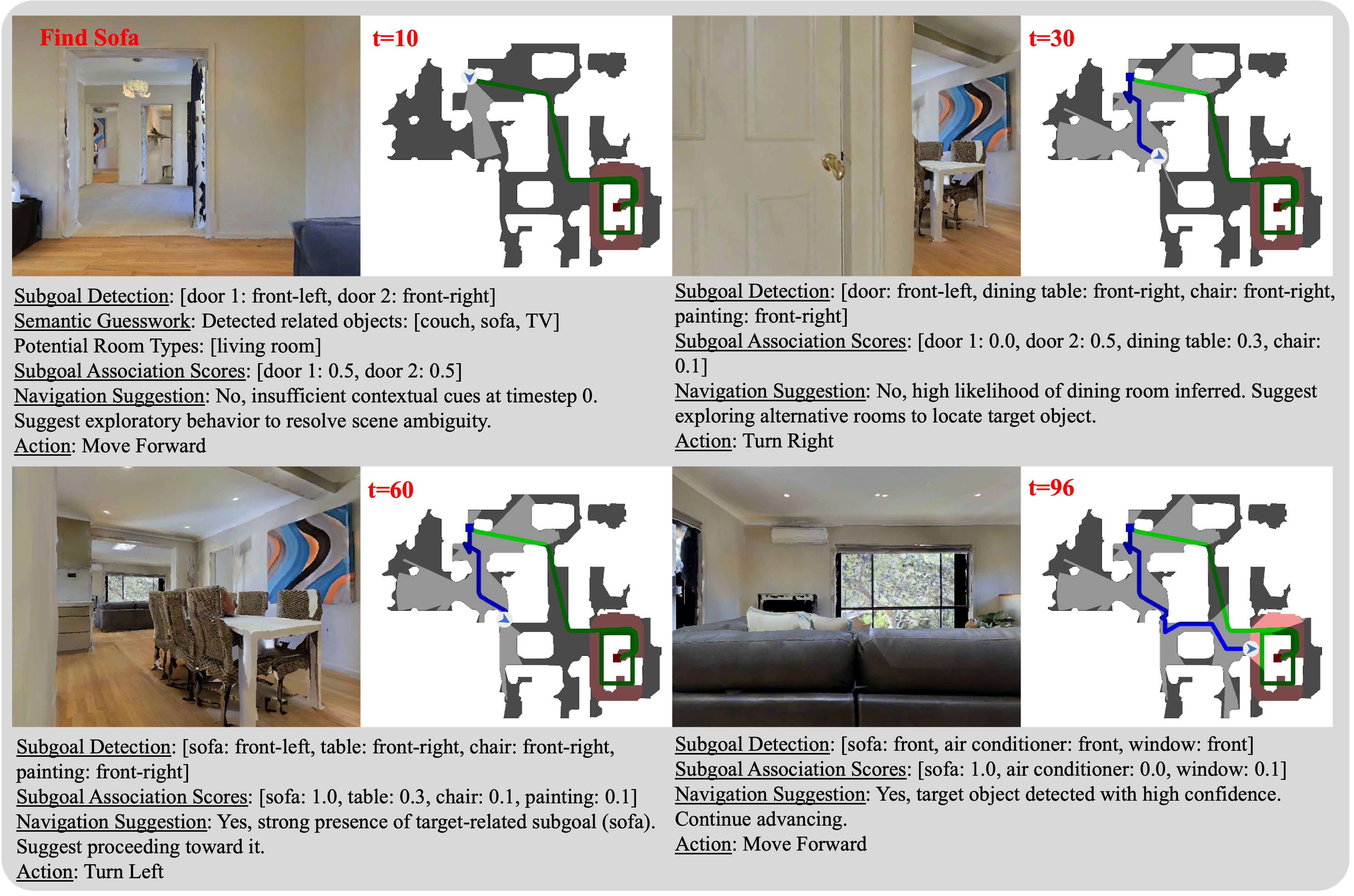}
    \caption{Zero-shot generalization results on MP3D Val. The figure illustrates how CL-CoTNav navigates in unseen scene layouts. The predicted navigation path is shown in blue and shortest path is shown in green. SPL = 0.71, ep\_length = 97.}
\end{figure*}

\subsection{Zero-Shot Generalization Test}

\begin{table}[t]
    \centering
    \caption{Zero-Shot ObjectNav Results on MP3D Val Split. We report Success Rate (SR) and Success weighted by Path Length (SPL) under two generalization settings: novel object goals and novel scenes. All models are trained on 35k trajectories.}
    \begin{tabular}{lcccc}
        \hline\hline
        \multirow{2}{*}{\textbf{Method}} & \multicolumn{2}{c}{\textbf{Novel Object Goals}} & \multicolumn{2}{c}{\textbf{Novel Scenes}} \\
        & SR (\%) & SPL (\%) & SR (\%) & SPL (\%) \\
        \hline
        Baseline~\cite{zhu2017target} & 22.7 & 5.1 & 25.3 & 6.1 \\
        Habitat-Web~\cite{ramrakhya2022habitat} & 26.5 & 7.4 & 28.2 & 9.0 \\
        VLFM~\cite{yokoyama2024vlfm} & 34.2 & 14.8 & 35.8 & 16.5 \\
        SSNet~\cite{zhao2023zero} & 30.2 & 10.8 & 31.1 & 12.1 \\
        DivScene~\cite{wang2024divscene} & 44.1 & 19.1 & 46.7 & 21.3 \\
        CL-CoTNav (35k) & \textbf{55.2} & \textbf{25.7} & \textbf{58.5} & \textbf{27.4} \\
        \hline\hline
    \end{tabular}
    \label{tab:zero_shot_results}
\end{table}

Table~\ref{tab:zero_shot_results} reports the performance of various models on zero-shot ObjectNav under two settings: novel object categories and novel scenes. CL-CoTNav achieves the highest success rate and SPL in both settings, demonstrating its strong generalization ability through hierarchical reasoning and confidence-aware learning.

For novel object generalization, models that incorporate semantic priors—such as VLFM, SSNet, DivScene, and CL-CoTNav—outperform traditional RL (Baseline) and IL (Habitat-Web). Although trained on expert demonstrations, Habitat-Web fails to generalize well to unseen objects, likely due to overfitting to category-specific patterns without broader semantic reasoning. VLFM benefits from frozen vision-language embeddings but lacks fine-tuning, resulting in low SPL due to inefficient trajectory planning. SSNet improves upon RL by modeling object-object associations but remains limited by its static representations. DivScene introduces chain-of-thought reasoning during VLM finetuning, leading to improved generalization, but is constrained by reliance on shortest-path supervision.
CL-CoTNav outperforms all baselines by explicitly modeling both object-level and room-level semantic relationships and refining decision-making through confidence-based weighting. This structured reasoning enables the agent to infer the likely location of unseen targets and navigate efficiently without memorized spatial priors.

In the novel scene generalization setting, the focus shifts to layout adaptation. RL-based methods struggle due to overfitting to seen environments. While SSNet introduces object-aware exploration cues, its performance remains limited in unfamiliar layouts. Imitation learning models, including Habitat-Web, experience significant performance drops, suggesting limited adaptability to new spatial arrangements. VLFM achieves better generalization due to its strong semantic priors, but its frozen architecture limits efficiency. DivScene benefits from intermediate reasoning but remains less effective than CL-CoTNav in navigating complex spatial layouts.
CL-CoTNav maintains a relatively small performance gap between training and zero-shot testing, indicating superior generalization. Its hierarchical planning allows flexible adaptation to novel configurations, while the closed-loop mechanism suppresses unreliable supervision, improving trajectory quality.

Overall, these results highlight the importance of structured multi-turn reasoning and adaptive learning in achieving robust generalization. CL-CoTNav bridges the limitations of prior approaches by integrating semantic reasoning with confidence-aware training, offering a scalable solution for zero-shot ObjectNav. Despite the model's fine-tuning on a 2B parameter scale, the multi-turn inference time remains at 1.2 seconds, making it suitable for practical applications. 
Future extensions may explore online adaptation to further enhance real-time deployment in unseen environments.

\subsection{Ablation Study}
This section isolates the contributions of hierarchical reasoning and closed-loop learning within the CL-CoTNav framework, analyzing their impact on zero-shot generalization.

As shown in Table~\ref{tab:ablation_study}, proposed H-CoT plays a central role in boosting generalization. Compared to a baseline using pure human annotations—where the QA format is limited to querying the target object and returning the human-demonstrated action—standard CoT prompting \cite{chen2024vision} improves performance by introducing intermediate reasoning. However, H-CoT further amplifies these gains by introducing a two-stage structure that separately models subgoal identification and semantic reasoning across room- and object-level contexts. This design leads to better-informed decision-making, significantly reducing failure cases arising from reactive or shallow policies.

Building on H-CoT, the Closed-Loop CoT mechanism provides an additional performance boost. By incorporating reasoning confidence scores into the training loop, the model learns to prioritize high-quality supervision while down-weighting noisy or semantically ambiguous samples. This adaptive loss weighting refines decision-making and enhances robustness under distribution shifts, particularly in unseen environments. The improvement from H-CoT to CL-CoTNav highlights the value of confidence-aware learning in mitigating the effect of hallucinated associations and unreliable intermediate predictions.

\begin{table}[h]
\caption{Ablation study results on MP3D Val (unseen scenes). This study evaluates the impact of hierarchical reasoning (H-CoT) and adaptive learning (Closed-Loop CoT) on generalization.}
\centering
\begin{tabular}{lcc}

\hline\hline
\textbf{Method} & \textbf{Success (↑)} & \textbf{SPL (↑)} \\
\hline
Pure Text (Human Annotations)  & 24.3\% & 6.5\%  \\
Standard CoT \cite{chen2024vision}  & 36.5\% & 15.8\% \\
H-CoT (Hierarchical CoT Only)  & 52.9\% & 23.1\% \\
CL-CoTNav (H-CoT + Closed-Loop) & 55.2\% & 25.7\% \\
\hline\hline
\end{tabular}

\label{tab:ablation_study}
\end{table}

Taken together, these results validate that CL-CoTNav’s superior generalization stems not only from its structured reasoning process, but also from its ability to selectively learn from trustworthy examples. The combined effect of hierarchical decomposition and closed-loop feedback forms a scalable framework for ObjectNav, enabling agents to move beyond passive imitation and toward actively adaptive navigation.

\section{Conclusion and Future Work}
In this work, we introduce Closed-Loop Hierarchical Chain-of-Thought Navigation (CL-CoTNav), a vision-language model (VLM)-driven ObjectNav framework that integrates structured reasoning and closed-loop feedback into navigation decision-making.
We fine-tune a VLM using multi-turn question-answering (QA) data derived from human demonstration trajectories. This structured dataset enables hierarchical Chain-of-Thought (H-CoT) prompting, iteratively extract compositional knowledge and provide auxiliary navigation guidance.
Additionally, we propose a Closed-Loop H-CoT mechanism that incorporates confidence scores into training to prioritize high-confidence data pairs, enhancing robustness against hallucinated or incorrect reasoning. Extensive experiments in the AI Habitat environment demonstrated that our method achieves superior generalization to unseen scenes and target objects, outperforming state-of-the-art approaches in success rate (SR) and success weighted by path length (SPL). 

Despite these advancements, our approach still relies on imitation learning, which is inherently limited by the quality and coverage of the dataset. To overcome this, future work will explore offline RL to enhance policy learning beyond supervised data constraints. Additionally, online fine-tuning will be investigated to further adapt navigation policies in real-time, enabling more effective exploration and dynamic decision-making. Further, we plan to conduct physical experiments to validate the approach in realistic settings. By integrating offline and online RL adaptation, we aim to bridge the gap between structured reasoning and real-world navigation challenges, ensuring more robust and scalable ObjectNav performance.

\bibliographystyle{ieeetr}

\bibliography{ref.bib}

\end{document}